\documentclass[conference]{IEEEtran}
\IEEEoverridecommandlockouts
\usepackage{cite}
\usepackage{amsmath,amssymb,amsfonts}
\usepackage{algorithmic}
\usepackage[linesnumbered]{algorithm2e}
\usepackage{graphicx}
\usepackage[marginal]{footmisc}

\usepackage{flushend,cuted}
\usepackage{subfigure}
\usepackage{multirow}
\usepackage{textcomp}
\usepackage{bbm}
\usepackage{xcolor}

\def\BibTeX{{\rm B\kern-.05em{\sc i\kern-.025em b}\kern-.08em
    T\kern-.1667em\lower.7ex\hbox{E}\kern-.125emX}}
    
\begin{document}

% Title.
% ------
\title{DSDRNet: Disentangling Representation and Reconstruct Network for Domain Generalization}
% combine Adapter and Prompt
% Single address.
% ---------------
\author{Juncheng Yang, Zuchao Li, Shuai Xie, Wei Yu, Shijun Li}
%Address and e-mail should NOT be added in the submission paper. They should be present only in the camera ready paper. 

% \author{
% \IEEEauthorblockN{Jianpeng Liao}
% \IEEEauthorblockA{
% \textit{South China University of Technology} \\
% Guangzhou, Guangdong, China\\
% sejianpengliao@mail.scut.edu.cn}
% \and
% \IEEEauthorblockN{Jun Yan}
% \IEEEauthorblockA{
% \textit{Concordia University} \\
% Montreal, Quebec, Canada\\
% jun.yan@concordia.ca}
% \and
% \IEEEauthorblockN{Qian Tao\IEEEauthorrefmark{1}}
% \IEEEcompsocitemizethanks{\IEEEauthorrefmark{1}Corresponding author.}
% \IEEEauthorblockA{
% \textit{South China University of Technology}\\
% Guangzhou, Guangdong, China\\
% \textit{Pazhou Lab}, Guangzhou, Guangdong, China\\
% taoqian@scut.edu.cn}
% }
\maketitle

\begin{abstract}
Domain generalization faces challenges due to the distribution shift between training and testing sets, and the presence of unseen target domains. Common solutions include domain alignment, meta-learning, data augmentation, or ensemble learning, all of which rely on domain labels or domain adversarial techniques. In this paper, we propose a Dual-Stream Separation and Reconstruction Network, dubbed DSDRNet. It is a disentanglement-reconstruction approach that integrates features of both inter-instance and intra-instance through dual-stream fusion. The method introduces novel supervised signals by combining inter-instance semantic distance and intra-instance similarity. Incorporating Adaptive Instance Normalization (AdaIN) into a two-stage cyclic reconstruction process enhances self-disentangled reconstruction signals to facilitate model convergence. Extensive experiments on four benchmark datasets demonstrate that DSDRNet outperforms other popular methods in terms of domain generalization capabilities.
\end{abstract}
\section{Introduction}
\label{sec:intro}

Currently, many deep learning tasks assume that the data distribution in the training set and the testing set is the same. However, real-world scenarios often deviate from this assumption, such as training a model on color images and testing it on black and white images. Traditional training methods perform poorly when dealing with data with significant distribution differences. Common solutions~\cite{zhou2022domain, wang2022generalizing, zhou2020learning, zhou2021domain, wang2022out, yang2021adversarial, guo2023domaindrop, jiang2022domain} involve leveraging data from multiple source domains to enable a trained model to generalize to different and unseen domains. Another challenge arises when dealing with scarce experimental data. Collecting a large amount of experimental data may not be feasible in certain scenarios, such as fall detection for the elderly and Person Re-ID~\cite{zhou2021learning, zhang2022adaptive, jiao2022dynamically, qi2022label}. These challenges have gained increasing attention from researchers, making the development of models with strong generalization capabilities a focal point in the research community.

\begin{figure}[t] 
	\centering
    \includegraphics[width=0.84\linewidth]{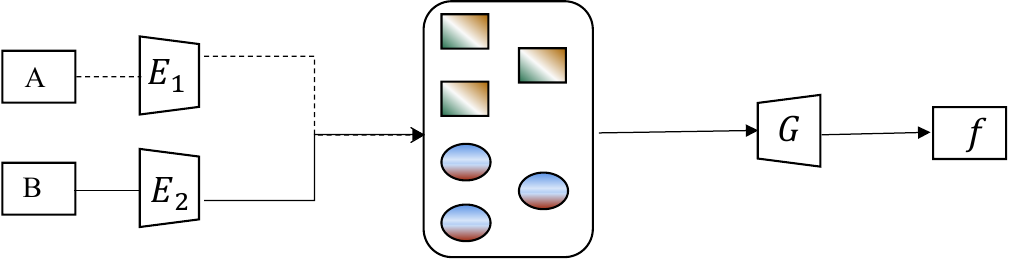}\\
    \small{(a) Traditional processing method~\cite{ganin2016domain,hoffman2018cycada}} \\
    \includegraphics[width=0.84\linewidth]{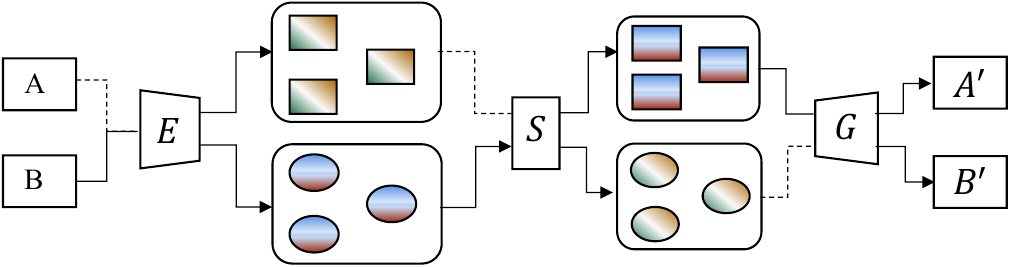}\\
    \small{(b) Disentangling representation and adaptation networks(DRANet)~\cite{lee2021dranet}}\\
    \includegraphics[width=0.84\linewidth]{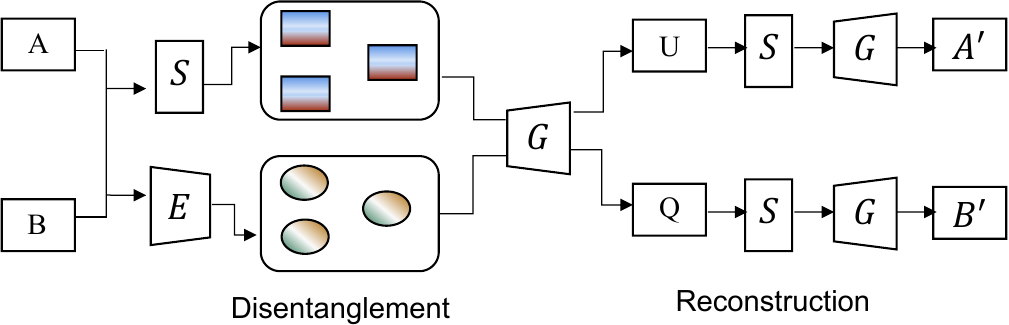}\\
    \small{(c) Our feature separation and domain generalization(\textit{DSDRNet})}\\
    \vspace{2mm}
	\caption{DSDRNet and Competing Methods: Domain Adaptation~\cite{ganin2016domain,hoffman2018cycada}, Representation Disentanglement~\cite{lee2021dranet}. DSDRNet learns a generalized feature representation through disentanglement, encoding, and reconstruction, consisting of the cyclic reconstruction loop with E, S, and G.}
	\label{fig:dual}
	\vspace{-3mm}
\end{figure}

Most existing domain generalization (DG) methods typically draw inspiration from domain adaptation (DA) approaches. The common practice is to perform feature-level alignment in the source domain, leveraging the powerful nonlinear fitting capability of deep networks to learn the invariance in the source domain. However, this approach does not necessarily enable the model to truly understand the features of images. Considering two images from different distributions, when using a feature extractor to generate semantics, a good feature extractor should achieve accurate semantics for each input image. So, how can we determine if the model truly understands an image? Disentanglement reconstruction is a common approach to addressing this issue. To ensure that the model, after disentangling and reconstructing the image, has stronger representation capabilities than before, this places more stringent requirements on deep networks.

According to this idea, this paper introduces a new disentanglement-reconstruction strategy, as illustrated in Figure~\ref{fig:dual}. Figure~\ref{fig:dual}(a) depicts a feature-based unsupervised cross-domain adaptation method~\cite{ganin2016domain,hoffman2018cycada}, which maps data from different domains to a common feature space. However, it loses some features due to enforcing space alignment, leading to a significant drop in model performance when dealing with unknown categories.
Figure~\ref{fig:dual}(b) disentangles images to the latent space using a linear disentanglement structure~\cite{zhang2018style} or untangles the images to the latent space, merging style factors into content factors to form new images~\cite{lee2021dranet}. Although the model understands the content of the images based on a unified encoder and disentanglement, it lacks generalization and performs poorly when dealing with new categories.

Traditional one-stage disentangle-reconstruct methods only treat AdaIN~\cite{huang2017arbitrary} as a style feature extractor without incorporating it into the image reconstruction and disentanglement processes. This inspired us to adopt a two-stage cycle reconstruction approach, as shown in Fig.~\ref{fig:dual}(c). This paper employs a dual-stream image blending reconstruction disentanglement method, providing strong supervision for AdaIN~\cite{huang2017arbitrary} to reconstruct its style, this encourages the model to learn style features with better disentangle-reconstruct capabilities.

We believe that a good disentangle model should not only possess the capability of two-stage cycle reconstruction but also have the property of maintaining almost unchanged semantic distances between reconstructed images. Based on this principle, we propose a dual-stream image blending reconstruction disentanglement method. During the model training process, we take full advantage of the characteristic that the blended reconstructed images and the original images exhibit approximate semantic space distances. In addition to using intra-instance similarity supervision, we also introduce inter-instance semantic distance supervision, allowing the model's reconstruction to maintain semantic space consistency and accurately locate and encode ground-truth semantics. Ultimately, this enables the model to learn generalization feature representations from different source domains, addressing the issue of performance degradation due to domain distribution differences.

We highlight the contributions of this work as follows:
\begin{itemize}

    \item This paper proposes a novel domain generalization method, which explores features from both within-instance and cross-instance perspectives in terms of image content and semantics. By leveraging the relationship between reconstructed images and original images, the proposed method enhances the model's generalization capabilities.
     \item We incorporate AdaIN into a two-stage cyclic reconstruction loop, providing stronger supervisory signals for image self-reconstruction.
    \item Combining with reconstruction, cross-cycle consistency, and semantics adversarial, DSDRNet ensures semantic consistency across different sources, achieving SOTA performance over prevailing baselines on four benchmarks.
\end{itemize} 

%------------------------------------------------------------------------
\section{Related Work}
\label{sec:formatting}
\subsection{Domain Generalization}
DG requires us to learn a model from a given number of source domains that can make good predictions about the unseen data distribution. MixStyle~\cite{zhou2020domain} introduced the mixing of source domains with a certain probability to generate new samples, aiming to enhance the model's generalization capability. UFDN~\cite{liu2018unified} proposed a unified feature disentangle for multi-domain image translation. MTAE~\cite{ghifary2015domain} provided a general framework for domain generalization based on multi-task learning. MMD-AAE~\cite{li2018domain} utilized maximum mean difference and adversarial auto-encoder for domain generalization. DIVA~\cite{ilse2020diva} divided the data into domain information, category information, and other information for the structure, and used VAE for data reconstruction to improve the generalization of the system. CCSA~\cite{motiian2017unified} proposed a unified architecture of siamese network to solve the generalization problem in the vision domain. CrossGrad~\cite{shankar2018generalizing} utilizes domain information to assist samples, using domain-guided data augmentation to improve the model's generalization. DDAIG~\cite{zhou2020deep} proposed to divide the model into label classifier, DoTNet, and domain classifier, using DoTNet to map source domain data to the unseen domain. L2A-OT~\cite{zhou2020learning} proposed to model the distribution between the synthesized pseudo-novel domain and the source domain, and maximize their divergence using the optimal transport. D-SAM~\cite{d2018domain} proposed a generic architecture that could improve the generalization of the system while maintaining the separation of existing source domain information and at the same time utilizing common information. DRANet~\cite{lee2021dranet} proposed a content-adaptive domain transfer network that helped to preserve the scene structure and transferring style. JiGen~\cite{carlucci2019domain} proposed an algorithm similar to solving a jigsaw puzzle to improve the generalization of the system. DAEL~\cite{zhou2021domain} utilized a deep adversarial auto-encoder to extract domain-specific features from class labels. 
Epi-FCR~\cite{li2019episodic} decomposed the deep network into feature extractor and classifier components and trained each component using the current domain partners.
% MetaReg~\cite{balaji2018metareg} proposed to learn a meta-learning framework to improve the system's generalizability. FIXED~\cite{lu2022fixed} enhanced the identification of domain-invariant features by enlarging the margin of the inter-class distance. 
\subsection{Disentangling Representations}
Disentangling representation was the analysis of potential generative factors inside the data from the perspective of data generation by studying the physical structure within the data. Bengio et al.~\cite{bengio2009learning} introduced a variance factor disentanglement problem. VAE~\cite{kingma2013auto} proposed modeling real data from a maximum likelihood perspective by calculating the KL divergence between the true posterior and the variational posterior to achieve data disentanglement.
${\beta}$-VAE~\cite{higgins2017beta} improved disentanglement performance by increasing the value of ${\beta}$, making the learned posterior distribution statistically similar to the prior distribution. AAE~\cite{makhzani2015adversarial} suggested using adversarial networks to measure the similarity between the posterior and prior distributions to enhance the representational capacity of latent variables. DRAW~\cite{gregor2015draw} introduced a deep recurrent network architecture to learn historical states for simple detection and object tracking. 
CDD~\cite{gonzalez2018image} leveraged a pair of generative adversarial networks to achieve a cross-domain bidirectional image translation model. 
MLVAE~\cite{bouchacourt2018multi} separated latent representations of grouped data-related factors by exchanging and sharing the bottom-level representations of samples.
MixNMatch~\cite{li2020mixnmatch} achieved a disentangled representation of pose, texture, and background for single-object scenes through data factorization and layered generation.

% \subsection{Mixing Method}
% Mixup~\cite{zhang2018mixup} was a simple data augmentation strategy that improves model generalization by performing a linear combination of the input data. Manifold Mixup~\cite{verma2019manifold} used semantic interpolation as an additional training signal to mixup two groups of hidden layers to make the decision boundary of the neural network smoother. AdaMixup~\cite{guo2019mixup} added a new classifier network during data mixing to address issues with mixed samples and data points. CutMix~\cite{yun2019cutmix} swapped regions between two images and weights the labels of the generated images to create new images. StyleMix~\cite{hong2021stylemix} was used for mixing style information to generate new images. StyleGAN~\cite{gowal2020achieving} utilized separated representations to generate image perturbations from the real world. Training the model to remain invariant to these perturbations enhances the system's robustness. PuzzleMix~\cite{kim2020puzzle} proposed optimizing mixup using salient information and statistical data from natural instances. AlignMixup~\cite{venkataramanan2022alignmixup} advocated for interpolating the local structure of the feature space to achieve semantic alignment. 

\begin{figure*}[!t]
\centering
\includegraphics[width=0.85\linewidth]{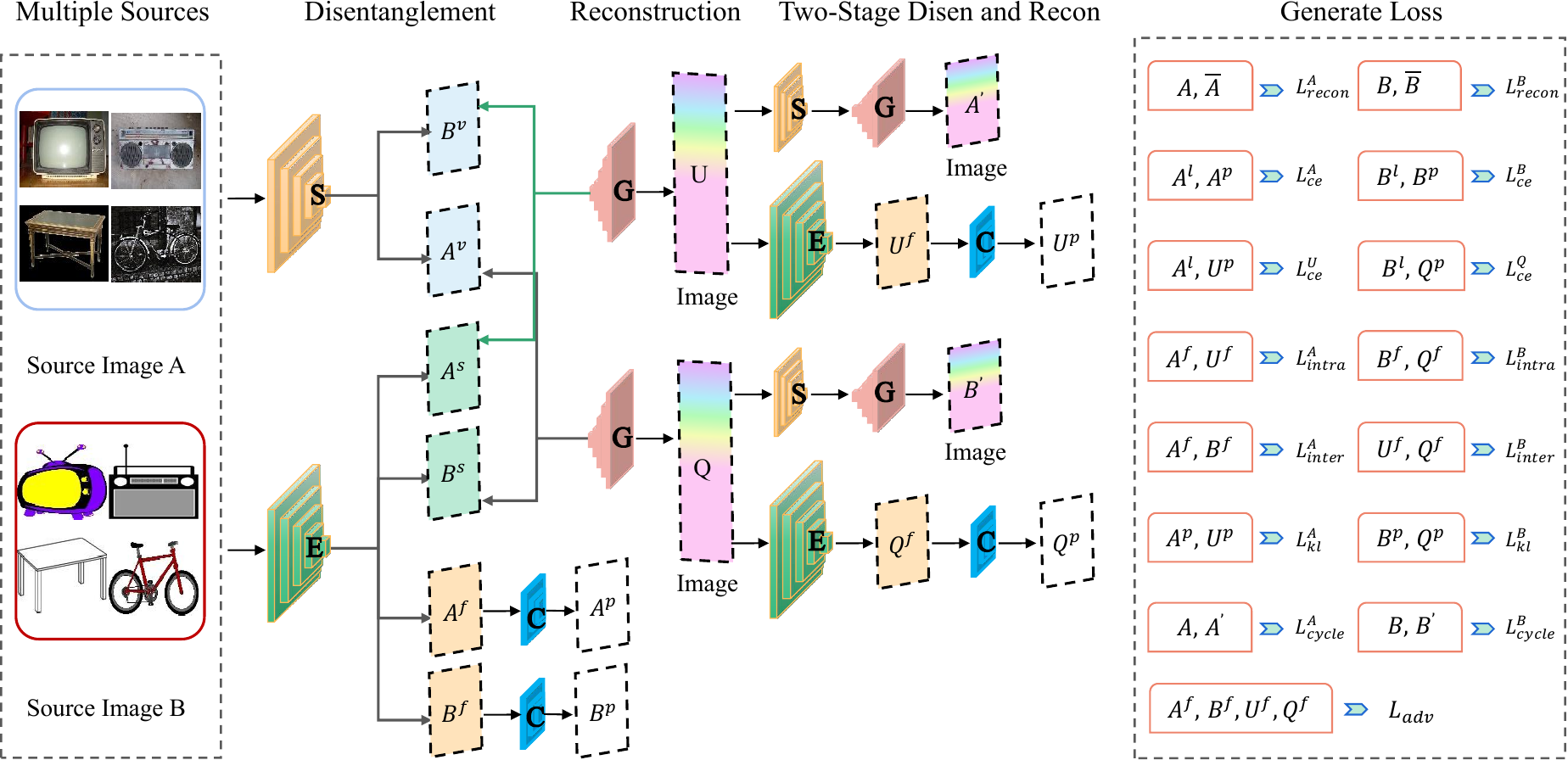}
% \caption{Overview of the proposed DSDRNet framework.}
\caption{Illustration of the proposed DSDRNet. Specifically, in the first stage, the disentanglement-reconstruction phase, we randomly select samples $A$ and $B$ from the source domain. These samples are processed through the separated $S$ and encoding $E$ modules, extracting attributes $A^v$, semantics $A^s$, and feature information $A^f$ from sample $A$, and attributes $B^v$, semantics $B^s$, and feature information $B^f$ from sample $B$. The generator $G$ utilizes ($A^s$, $B^v$) and ($B^s$, $A^v$) to generate new images $U$ and $Q$ , and utilizes ($A^s$, $A^v$) and ($B^s$, $B^v$) to reconstruct images $\bar A$ and $\bar B$. In the second stage, $U$ and $Q$ are further processed through $S$, $E$, and $G$ to generate new images $A'$ and $B'$. The model improves generalization performance through reconstruction $\mathcal{L}_{\text{recon}}$, intra-instance reconstruction $\mathcal{L}_{\text{intra}}$, inter-instance reconstruction $\mathcal{L}_{\text{inter}}$, cross-cycle consistency $\mathcal{L}_{\text{cycle}}$, classification $\mathcal{L}_{\text{ce}}$, $\mathcal{L}_{\text{kl}}$, and semantic adversarial $\mathcal{L}_{\text{adv}}$ terms.}
\label{fig:framework}
\end{figure*}

\section{Method}
\subsection{Problem Definition}
Assuming there are M source domains ${Source} =\{ {{Source}_1},\dots,{{Source}_M}\}$, where each source domain ${Source}_m = \{{x_i,y_i}\}_{i=1}^N$ contains N labeled samples, the goal of the model is to learn a powerful and generalizable predictive function $h$ from the data of M source domains. The objective is to minimize the error on the unseen target domain ${T}$: ${min}\mathbb{E}{(x,y)\mathcal{\in}T}[l(h(x),y)]$.

\subsection{Proposed DSDRNet}

The main process of DSDRNet is depicted in Figure~\ref{fig:framework}, which includes an encoder $E$, a disentangle $S$, used in this paper as AdaIN~\cite{huang2017arbitrary}, a generator $G$, a discriminator $D$, and a classifier $C$. DSDRNet aims to disentangle the semantics and attributes of samples into different latent spaces through a dual-stream network, training a highly robust and generalizable classifier. It utilizes semantic distance supervision to constrain the similarity of images and reconstructs the semantics and attributes of samples. Through adversarial training and consistency principles, it enables flexible manipulation and augmentation of the training data. Ultimately, this approach equips the model with excellent generalization capabilities for unseen samples.
%Images $A$ and $B$ are randomly selected samples from the source domain. $A^l$ and $B^l$ represent the labels of samples $A$ and $B$, respectively. The samples are passed through the encoder $E$ and the disentangle $S$ to extract semantic and attribute information. $A^v$ represents the attribute vector of sample $A$ disentangled by $S$, and $B^v$ represents the attribute vector of sample $B$ disentangled by $S$. Similarly, $A^s$ represents the semantic vector extracted from sample $A$ by $E$, and $B^s$ represents the semantic vector extracted from sample $B$ by $E$. Additionally, $A^f$ represents the feature vector extracted from sample $A$ by $E$, and $B^f$ represents the feature vector extracted from sample $B$ by $E$. Moreover, $A^s$ and $B^v$ are used by $G$ to generate the mixed image $U$, and $B^s$ and $A^v$ are used by $G$ to generate the mixed image $Q$. 

% Similarly, $U^v$ denotes the attribute vector of sample $U$ disentangled by $S$ and $Q^v$ denotes the attribute vector of sample $Q$ disentangled by $S$. $U^s$ denotes the semantic vector extracted from sample $U$ via $E$ and $Q^s$ denotes the semantic vector extracted from sample $Q$ via $E$. In addition, $U^f$ denotes the feature vector extracted from sample $U$ via $E$ and $Q^f$ denotes the feature vector extracted from sample $Q$ via $E$. The generator $G$ generates the reconstructed image $A'$ using the semantic vector $U^s$ and the attribute vector $Q^v$, and generates the reconstructed image $B'$ using the semantic vector $Q^s$ and the attribute vector $U^v$.

\textbf{Disentangling Representation.}$\quad$
The model's objective is to reduce semantic differences among generated samples belonging to the same class. Unlike traditional approaches that use a unified encoder and decoder for feature extraction, this paper employs two separate networks: the disentangle $S$ and the encoder $E$. They independently perform the disentangling operation on samples $A$ and $B$ at the pixel level, resulting in $A^v$, $A^s$, $A^f$, $B^v$, $B^s$, and $B^f$, as shown below:
\begin{equation}
\begin{aligned}
    &\{A^v,B^v\} = \{{S(A), S(B)}\},\\
    &\{A^s,B^s,A^f,B^f\} = \{{E(A), E(B)}\},
    % &\{A^f,B^f\} = \{{E_A(A), E_B(B)}\},
\end{aligned}
\end{equation}
where $A^v$ represents the attributes value of sample $A$ extracted by the disentangle $S$, while $B^v$ represents the attributes of sample $B$ extracted by the disentangle $S$, $A^s$ and $A^f$ denote the semantic and feature information of sample $A$ obtained through the encoder $E$, respectively. Similarly, $B^s$ and $B^f$ refer to the semantic and feature information of sample $B$ acquired through the encoder $E$.

\textbf{Reconstruction.}$\quad$
After the disentanglement operations in the previous stage, we obtained $A^v$, $A^s$, $A^f$, $B^v$, $B^s$, and $B^f$. We utilize the generator $G$ to generate reconstructed images $\bar{A}$ with the semantics $A^s$ and attributes $A^v$ of sample $A$, as well as $\bar{B}$ with the semantics $B^s$ and attributes $B^v$ of sample $B$. The results are as follows: 
\begin{equation}
\begin{aligned}
    \bar{A} = G(A^s,A^v),\\
    \bar{B} = G(B^s,B^v).
\end{aligned}
\end{equation}
At the same time, we utilize the generator $G$ to generate reconstruction images $U$ with the semantic $A^s$ of sample $A$ and the attribute $B^v$ of sample $B$, as well as reconstruction images $Q$ with the semantic $B^s$ of sample $B$ and the attribute $A^v$ of sample $A$. The results are as follows:
\begin{equation}
\begin{aligned}
    U = G(A^s,B^v),\\
    Q = G(B^s,A^v),
\end{aligned}
\end{equation}
where $U$ and $Q$ refer to the newly generated images.

We generate images $U$ and $Q$ through the first reconstruct operation. Then, we operate on $U$ using the disentangle $S$ to obtain attribute values $U^v$. Similarly, by using the encoder $E$, we obtain the semantic value $U^s$ and feature value $U^f$ of $U$. Likewise, we can obtain the attribute values $Q^v$, semantic value $Q^s$, and feature value $Q^f$ of $Q$. Using these semantics and attributes, we perform another reconstruct operation to obtain images $A'$ and $B'$. The specific process is as follows:
\begin{equation}
\begin{aligned}
    A' = G(U^s,Q^v),\\
    B' = G(Q^s,U^v),
\end{aligned}
\end{equation}
where $U^s$ and $U^v$ represent the semantics and attributes of $U$, respectively. Similarly, $Q^s$ and $Q^v$ represent the semantics and attributes of $Q$. $A'$ and $B'$ denote the newly generated images after the disentangling and reconstruction operations.

\subsection{Training Loss}
The overall loss of our framework consists of the 
intra-instance reconstruction loss $\mathcal{L}_{\text{intra}}$,
inter-instance reconstruction loss $\mathcal{L}_{\text{inter}}$,
self-reconstruct loss $\mathcal{L}_{\text{recon}}$,
cross-cycle consistency loss $\mathcal{L}_{\text{cycle}}$,
semantics adversarial loss $\mathcal{L}_{\text{adv}}$,
classification loss $\mathcal{L}_{\text{kl}}$ and $\mathcal{L}_{\text{ce}}$,
% mutual information minimization loss $\mathcal{L}_{mine}$,   
and with the balancing term $\alpha$:
\begin{equation}
\begin{aligned}
\mathcal{L}&=\alpha_1\mathcal{L}_{\text{intra}}
+\alpha_2\mathcal{L}_{\text{inter}}+\alpha_3 \mathcal{L}_{\text{recon}}\\
&+\alpha_4\mathcal{L}_{\text{cycle}}+\alpha_5\mathcal{L}_{\text{adv}}
+\alpha_6\mathcal{L}_{\text{kl}} +\alpha_7\mathcal{L}_{\text{ce}}.
\end{aligned}
\end{equation}

\textbf{Intra-instance Reconstruction Loss.}$\quad$
To ensure consistency between the features of the original image $A^f$ and the recombined features $U^f$ in the feature space, we need to calculate the angle between the two vectors, $A^f$ and $U^f$, to measure their similarity. Cosine similarity is well-suited for such calculations, as its values range from -1 to 1, with values closer to 1 indicating higher similarity. Therefore, the final intra-instance reconstruction loss can be expressed as:
\begin{equation}
    \mathcal{L}_{\text{intra}}^A = \frac{1}{N}\sum_{i=1}^N(1-\frac{A_i^f \cdot U_i^f}{\|A_i^f\| \cdot\|U_i^f\|}).
\label{eq:intra_loss}
\end{equation}
This loss function quantifies the similarity between the original and recombined features by measuring the cosine of the angle between the two feature vectors, intending to make them as similar as possible.

Similarly, we can obtain the cosine similarity loss between $B^f$ and $Q^f$ as  $\mathcal{L}_{\text{intra}}^B$. The overall intra-instance reconstruction loss can be represented as:
\begin{equation}
    \mathcal{L}_{\text{intra}} = \mathcal{L}_{\text{intra}}^A + \mathcal{L}_{\text{intra}}^B.
\label{eq:intra_instan_loss}
\end{equation}
The cosine similarity loss allows us to assess the similarity between the two feature vectors by quantifying their angular difference. By minimizing this loss, we aim to ensure that the reconstructed images remain similar to the original images in terms of feature representations.

\textbf{Inter-instance Reconstruction Loss.}$\quad$
Inter-instance reconstruction refers to the measurement of the dissimilarity between the reconstructed images of different instances. It captures the discrepancy between the features and attributes of different instances in the reconstructed space. The inter-instance reconstruction loss can be calculated using various metrics, such as Euclidean distance, L1 distance, or Structural Similarity Index Measure (SSIM). The purpose of the inter-instance reconstruction loss is to encourage the generated images to have distinct attributes and avoid blending or mixing the attributes of different instances in the reconstruction process.

To further constrain the similarity of image reconstruction, we utilize the L1 norm to enforce cosine similarity loss between them. Given the feature information of two images $A$ and $B$ as $A^f$ and $B^f$ respectively, and the feature information of two distinct images $U$ and $Q$ as $U^f$ and $Q^f$, the inter-instance reconstruction loss can be expressed as:
\begin{equation}
    \mathcal{L}_{\text{inter}} = \lVert \text{cos}(A^f, B^f), \text{cos}(U^f, Q^f) \lVert_{1},
\label{eq:inter_loss}
\end{equation}
where $\text{cos}(A^f, B^f)$ presents the cosine similarity loss between $A^f$ and $B^f$, $\text{cos}(U^f, Q^f)$ presents the cosine similarity loss between $U^f$ and $Q^f$.  

\textbf{Self-reconstruction Loss.}$\quad$
We apply an L1 loss to train $E$, $S$, and $G$, reducing the disparity between the input images $A$, $B$ and their corresponding reconstructed images $\bar{A}$, $\bar{B}$, the reconstructed loss is defined as:
\begin{equation}
    \begin{aligned}
        &\mathcal{L}_{\text{recon}}^A=\lVert A - \bar{A} \lVert_{1},\\
        &\mathcal{L}_{\text{recon}}^B=\lVert B - \bar{B} \lVert_{1}, \\
        &\mathcal{L}_{\text{recon}}= \mathcal{L}_{\text{recon}}^A + \mathcal{L}_{\text{recon}}^B,
    \end{aligned}
\label{recon_loss}
\end{equation}
where $\bar{A} = G(A^s,A^v)$ and $\bar{B} = G(B^s,B^v)$, respectively.

\textbf{Cross-cycle Consistency Loss.}$\quad$
Consistency refers to maintaining invariant between perturbed images and features to learn robust representations. In this paper, consistency loss attempts to maintain composition invariant of semantics and attributes when the domain-transferred image is re-projected into the representation space. After two translation stages, the reconstructed image should exhibit consistency in semantics and attributes with the original image. To enforce this constraint, we define the cross-cycle consistency loss as follows:

\begin{equation}
    \begin{aligned}
        &\mathcal{L}_{\text{cycle}}^A=\lVert A - A' \lVert_{1},\\
        &\mathcal{L}_{\text{cycle}}^B=\lVert B - B' \lVert_{1}, \\
        &\mathcal{L}_{\text{cycle}}= \mathcal{L}_{\text{cycle}}^A + \mathcal{L}_{\text{cycle}}^B,
    \end{aligned}
\label{cyc_loss}
\end{equation}
where $A'=G(U^v, Q^s)$ and $B'=G(U^s, Q^v )$.
This loss explicitly encourages consistency in semantics and the appearance of attributes, which is more conducive to improving the generalization of the model.

\textbf{Semantics Adversarial Loss.}$\quad$
To ensure semantic consistency between the original image $A$ and the synthesized image $U$, we introduce a content discriminator $D$ to distinguish the differences between real images and synthesized images. Additionally, we employ the generator $G$ to attempt image generation. The same approach is applied to supervise semantic consistency between the original image $B$ and the synthesized image $Q$, the semantics-based adversarial loss can be expressed as:

\begin{equation}
\begin{aligned}
    \mathcal{L}_{\text{adv}} = \mathbb{E}_{A}[\text{log}{D(A^f)}+ \text{log}{(1-D(A^f)}]\\  
  + \mathbb{E}_{B}[\text{log}{D(B^f)}+ \text{log}{(1-D(B^f)}] \\
  + \mathbb{E}_{U}[\text{log}{D(U^f)}+ \text{log}{(1-D(U^f)}] \\
  + \mathbb{E}_{Q}[\text{log}{D(Q^f)}+ \text{log}{(1-D(Q^f)}]. 
\end{aligned}
\label{cont_loss}
\end{equation}

\textbf{Kullback-Leibler Divergence Loss.}$\quad$
To further ensure that the model maintains semantic invariant during reconstruction, we also apply a model distribution fitting loss to the classification predictions of the original image $A$, denoted as $A^p$, and the classification predictions of the reconstructed image $U$, denoted as $U^p$, the results are as follows:
\begin{equation}
    \mathcal{L}_{kl}^A = \frac{1}{K}\sum_{k=1}^K P(A_k^p)\text{log}\frac{P(A_k^p)}{P(U_k^p)},
\label{eq:kl_loss}
\end{equation}
where $P(A_k^p)$ represents the probability distribution of $A_k^p$, $P(U_k^p)$ represents the probability distribution of $U_k^p$.
Similarly, the Kullback-Leibler Divergence Loss $\mathcal{L}_{\text{kl}}^B$ can be obtained for $B_k^p$ and $Q_k^p$ over the probability distribution, and the final result can be expressed as:
\begin{equation}
    \mathcal{L}_{\text{kl}} = \mathcal{L}_{\text{kl}}^A + \mathcal{L}_{\text{kl}}^B.
\label{eq:final_kl_loss}
\end{equation}

\textbf{Classification Loss.}$\quad$
To improve the classification accuracy of the model, we employ multiple class identifiers, denoted as $C$ to enhance classification accuracy. The procedure is as follows, given samples $A$ and $B$, encoded by the encoder $E$ to obtain $A^f$ and $B^f$, they pass through the classifier $C$ to produce $A^p$ and $B^p$. The K-way class identifier $C$ is utilized for accurate label predictions, supervised by the cross-entropy loss:
\begin{equation}
    \mathcal{L}_{\text{ce}}^{A} = \frac{1}{K}\sum_{k=1}^K\mathbbm{1}[k=A_k^l]\text{log}(C(A_k^f)).
\label{eq:A_ce_loss}
\end{equation}

Similarly, the same procedure can be applied to obtain $\mathcal{L}_{\text{ce}}^{B}$, $\mathcal{L}_{\text{ce}}^{U}$, $\mathcal{L}_{\text{ce}}^{Q}$, the final overall cross-entropy loss can be expressed as:
\begin{equation}
    \mathcal{L}_{\text{ce}} = \mathcal{L}_{\text{ce}}^{A} + \mathcal{L}_{\text{ce}}^{B} +\mathcal{L}_{\text{ce}}^{U} +\mathcal{L}_{\text{ce}}^{Q}.
\label{eq:final_loss}
\end{equation}

\section{Experiment}
% APPENDIX部分的\label{appen:datasets}有关于数据集的详细介绍，\label{appen:method}部分有关于对比方法的介绍，\label{appen:exper_detail}部分有关于实验细节的介绍。
For a fair comparison with prior work, we select one domain as the test domain and utilize the remaining domains as source domains. To ensure the reliability of the results, we conducted three runs with different random seeds, averaging the top-1 classification accuracy as the performance measure and reporting the mean accuracy. 
In the~\ref{appen:datasets} section of the \textbf{APPENDIX}, detailed information about the datasets is provided. The~\ref{appen:methods} section introduces the comparative methods, while the~\ref{appen:exper_detail} section covers the experimental details.

\subsection{Experiments on Digits-DG}

\begin{table}
\centering
\caption{The testing performance of DSDRNet on the Digits-DG dataset.}
\label{tab:exp_Digits-DG}
\resizebox{0.45 \textwidth}{!}{
\begin{tabular}{l| c c c c |c}
%\begin{tabular}{p{1.9cm} p{1.2cm} p{1.5cm} p{1.2cm} p{1.2cm} p{1.1cm}}
\hline
Method & MNIST & MNISTM & SVHN & SYN & Avg \\ \hline
ERM & 95.8 & 58.8 & 61.7 & 78.6 & 73.7 \\
DANN~\cite{ganin2016domain} & 97.8 & 55.6 & 61.9 & 89.4 & 76.2 \\
CORAL~\cite{sun2016deep} & 97.6 & 57.7 & 57.8 & 90.1 & 75.8 \\
Mixup~\cite{zhang2018mixup} & 97.5 & 58.0 & 54.8 & 89.8 & 75.0 \\
Jigen~\cite{carlucci2019domain} & 96.5 & 61.4 & 63.7 & 74.0 & 73.9 \\
GroupDRO~\cite{sagawa2019distributionally} & 97.5 & 53.5 & 55.6 & \textbf{92.2} & 74.7 \\
RSC~\cite{huang2020self} & 97.8 & 56.3 & 62.4 & 89.3 & 76.4 \\
MixStyle~\cite{zhou2021domain} & 96.5 & 63.5 & 64.7 & 81.2 & 76.5 \\
ANDMask~\cite{parascandolo2020learning}  & 96.9 & 56.0 & 59.5 & 88.2 & 75.1 \\
CCSA~\cite{motiian2017unified} & 95.2 & 58.2 & {65.5} & 79.1 & 74.5 \\
CrossGrad~\cite{shankar2018generalizing} & {96.7} & {61.1} & 65.3 & {80.2} & {75.8} \\
DDAIG~\cite{zhou2020deep}  & {96.6} & {64.1} & {68.6} & {81.0} & {77.6} \\
MMD-AAE~\cite{li2018domain} & 96.5 & 58.4 & 65.0 & 78.4 & 74.6 \\
L2A-OT~\cite{zhou2020learning}  & {96.7} & {63.9} & {68.6} & {83.2} & {78.1} \\
\hline
DSDRNet                         & \textbf{98.4} & \textbf{65.2} & \textbf{70.1} & 89.8 & \textbf{80.9} \\
\hline
\end{tabular}}
\vspace{-0.1cm}
\end{table}

% \textbf{Results and Analysis.}$\quad$
As shown in Table~\ref{tab:exp_Digits-DG}, the DSDRNet achieved an accuracy of 98.4\% on the MNIST dataset, surpassing the ERM baseline by +2.6\% and outperforming the best DANN~\cite{ganin2016domain} and RSC~\cite{huang2020self} by +0.6\%. On the MNISTM dataset, the model achieved an accuracy of 65.2\%, surpassing ERM by +6.4\% and outperforming the best DDAIG~\cite{zhou2020deep} by +1.1\%. On the SVHN dataset, the model's performance exceeded ERM by +8.4\% and outperformed the current best DDAIG~\cite{zhou2020deep} and L2A-OT~\cite{zhou2020learning} by +1.5\%. On the SYN dataset, the model outperformed ERM by +11.2\%, but it scored lower than the best GroupDRO~\cite{sagawa2019distributionally} by 2.4\%. This difference can be attributed to the fact that the SYN dataset contains randomly inserted backgrounds, leading to more pronounced model oscillations. Overall, the DSDRNet achieved an average accuracy of 80.9\% across the four datasets, surpassing ERM by +7.2\% and outperforming the best L2A-OT~\cite{zhou2020learning} by +2.8\%.

\begin{table}[t]
\centering
\caption{The testing performance of DSDRNet on the PACS dataset.}
\label{tab:exp_PACS}
\resizebox{0.45 \textwidth}{!}{
\begin{tabular}{l| c c c c |c}
\hline
Method & Art & Cartoon & Photo & Sketch & Avg \\ \hline
ERM & 77.0 & 75.9 & 96.0 & 69.2 & 79.5  \\
DANN~\cite{ganin2016domain} & 78.7 & 75.3 & 94.0 & 77.8 & 81.4  \\
CORAL~\cite{sun2016deep} & 77.8 & 77.1 & 92.6 & 80.6 & 82.0  \\
Mixup~\cite{zhang2018mixup} & 79.1 & 73.5 & 94.5 & 76.7 & 80.9  \\
GroupDRO~\cite{sagawa2019distributionally} & 76.0 & 76.1 & 91.2 & 79.1 & 80.6  \\
RSC~\cite{huang2020self} & 79.7 & 76.1 & 95.6 & 76.6 & 82.0  \\
MMLD~\cite{matsuura2020domain} & 81.3 & 77.2 & 96.1 & 72.3 & 81.7  \\
ANDMask~\cite{parascandolo2020learning} & 76.2 & 73.8 & 91.6 & 78.1 & 79.9  \\
MixStyle~\cite{zhou2021domain} & 84.1 & \textbf{78.8} & 96.1 & 75.9 & 83.7  \\
SagNet~\cite{nam2021reducing} & 83.6 & 77.7 & 95.5 & 76.3 & 83.3  \\
CCSA~\cite{motiian2017unified} & 80.5 & 76.9 & 93.6 & 66.8 & 79.5 \\
CrossGrad~\cite{shankar2018generalizing} & 79.8 & 76.8 & 96.0 & 70.2 & 80.7 \\
DDAIG~\cite{zhou2020deep} & 84.2 & 78.1 & 95.3 & 74.7 & 83.1 \\
MMD-AAE~\cite{li2018domain} & 75.2 & 72.7 & 96.0 & 64.2 & 77.0 \\
L2A-OT~\cite{zhou2020learning} & 83.3 & 78.2 & 96.2 & 73.6 & 82.8 \\
D-SAM~\cite{d2018domain} & 77.3 & 72.4 & 95.3 & 77.8 & 80.7 \\
JiGen~\cite{carlucci2019domain} & 79.4 & 75.3 & 96.0 & 71.6 & 80.6 \\
Epi-FCR~\cite{li2019episodic} & 82.1 & 77.0 & 93.9 & 73.0 & 81.5 \\
DAEL~\cite{zhou2021domain} & 84.6 & 74.4 & 95.6 & 78.9 & 83.4 \\
MetaReg~\cite{balaji2018metareg} & 83.7 & 77.2 & 95.5 & 70.3 & 81.7 \\ 
\hline
DSDRNet & \textbf{85.2} & 78.4 & \textbf{96.8} & \textbf {81.2} & \textbf{85.4} \\
\hline
\end{tabular}}
\vspace{-0.1cm}
\end{table}

\subsection{Experiments on PACS}
Based on Table~\ref{tab:exp_PACS}, in the Art domain, DSDRNet achieves an accuracy of 85.2\%, surpassing ERM by 8.2\% and outperforming the best-performing method, DAEL~\cite{zhou2021domain}, by +0.6\%. In the Cartoon domain, the model achieves an accuracy of 78.4\%, outperforming ERM by +2.5\% and slightly below the best method, MixStyle~\cite{zhou2021domain}, by +0.4\%. In the Photo domain, DSDRNet outperforms ERM by +0.8\% and is +0.6\% better than the current best, L2A-OT~\cite{zhou2020learning}. In the Sketch domain, the model achieves an accuracy of 81.2\%, surpassing ERM by +12.0\% and outperforming the current SOTA CORAL~\cite{sun2016deep} by +0.6\%. On the PACS dataset as a whole, DSDRNet reaches an accuracy of 85.4\%, significantly outperforming similar comparison methods, exceeding ERM by +5.9\%, and outperforming MixStyle by +1.7\%.

\subsection{Experiments on OfficeHome}
As shown in Table~\ref{tab:exp_OfficeHome}, DSDRNet achieves an accuracy of 67.3\% on the OfficeHome dataset, surpassing ERM by +2.6\%, and outperforming the current best method, DAEL~\cite{zhou2021domain}, by +1.2\%. Specifically, in the Art domain, the model achieves an accuracy of 61.7\%, surpassing ERM by +2.8\%, and outperforming L2A-OT~\cite{zhou2020learning} by +1.1\%. In the Clipart domain, DSDRNet reaches 54.6\%, surpassing ERM by +5.2\%, and slightly below DAEL~\cite{zhou2021domain} by 0.5\%. In the Product domain, the model outperforms ERM by +0.9\% and surpasses L2A-OT~\cite{zhou2020learning} by +0.4\%. In the RealWorld domain, the model outperforms ERM by +1.6\% and surpasses L2A-OT~\cite{zhou2020learning} by +0.8\%. These results showcase the efficacy of DSDRNet in enhancing domain generalization performance on the OfficeHome dataset.

\begin{table}[t]
\centering
\caption{The testing performance of DSDRNet on OfficeHome dataset.}
\label{tab:exp_OfficeHome}
\resizebox{0.45 \textwidth}{!}{
\begin{tabular}{l| c c c c |c}
\hline
Method & Art & Clipart & Product & RealWorld & Avg \\ \hline
ERM & 58.9 & 49.4 & 74.3 & 76.2 & 64.7  \\
DANN~\cite{ganin2016domain} & 57.7 & 44.4 & 72.0 & 72.5 & 61.7  \\
CORAL~\cite{sun2016deep} & 58.8 & 48.8 & 72.3 & 73.6 & 63.4  \\
Mixup~\cite{zhang2018mixup} & 55.8 & 47.9 & 72.0 & 72.8 & 62.1  \\
GroupDRO~\cite{sagawa2019distributionally} & 57.6 & 48.8 & 71.5 & 73.2 & 62.8  \\
RSC~\cite{huang2020self} & 59.0 & 49.2 & 72.5 & 74.2 & 63.7  \\
ANDMask~\cite{parascandolo2020learning} & 56.7 & 45.9 & 70.7 & 73.2 & 61.6  \\
SagNet~\cite{nam2021reducing} & 60.2 & 45.4 & 70.4 & 73.4 & 62.3  \\
CCSA~\cite{motiian2017unified} & 59.9 & 49.9 & 74.1 & 75.7 & 64.9 \\
CrossGrad~\cite{shankar2018generalizing} & 58.4 & 49.4 & 73.9 & 75.8 & 64.4 \\
DDAIG~\cite{zhou2020deep} & 59.2 & 52.3 & 74.6 & 76.0 & 65.5 \\
MMD-AAE~\cite{li2018domain} & 56.5 & 47.3 & 72.1 & 74.8 & 62.7  \\
L2A-OT~\cite{zhou2020learning} & 60.6 & 50.1 & 74.8 & 77.0 & 65.6 \\
D-SAM~\cite{d2018domain} & 58.0 & 44.4 & 69.2 & 71.5 & 60.8 \\
JiGen~\cite{carlucci2019domain} & 53.0 & 47.5 & 71.5 & 72.8 & 61.2 \\
DAEL~\cite{zhou2021domain} & 59.4 & \textbf{55.1} & 74.0 & 75.7 & 66.1 \\
\hline
DSDRNet & \textbf{61.7} & 54.6 & \textbf{75.2} & \textbf{77.8} & \textbf{67.3} \\
\hline
\end{tabular}
}
\vspace{-0.1cm}
\end{table}

\subsection{Experiments on DomainNet}
As shown in Table~\ref{tab:exp_DomainNet}, on the DomainNet dataset, ERM achieves an average accuracy of 41.6\% across the 6 domains. In contrast, DSDRNet achieves an accuracy of 43.3\%, surpassing ERM by +1.7\% and outperforming MLDG~\cite{li2018learning} by +0.8\%. Specifically, in the Clipart domain, the model achieves an accuracy of 60.7\%, surpassing ERM by +2.3\% and outperforming MLDG~\cite{li2018learning} by +1.4\%. In the Infograph domain, DSDRNet reaches an accuracy of 21.5\%, surpassing ERM by +1.7\% and outperforming CORAL by +0.7\%. The model achieved an accuracy of 47.9\% in the Painting domain, which is slightly higher than ERM by +0.6\% and slightly lower than MLDG~\cite{li2018learning} by 0.9\%. The model is +2.4\% higher than ERM and +1.8\% higher than MLDG~\cite{li2018learning} on the Quickdraw domain. In the Real domain, the model performed +1.5\% better than ERM and +1.0\% better than MLDG~\cite{li2018learning}. The model is +2.0\% higher than ERM and +0.7\% higher than MLDG~\cite{li2018learning} on the Sketch domain. These results indicate that DSDRNet maintains a relatively high level of generalization when encountering datasets with a large number of samples and complex variations.
\begin{table}[t]
\centering
\caption{The testing performance of DSDRNet on the DomainNet dataset.}
\label{tab:exp_DomainNet}
\resizebox{0.48 \textwidth}{!}{
\begin{tabular}{l| c c c c c c |c}
\hline
Method & Clipart & Infograph & Painting & Quickdraw & Real & Sketch & Avg \\ \hline
ERM & 58.4 & 19.8 & 47.3 & 13.4 & 60.7 & 49.9 & 41.6  \\
IRM~\cite{arjovsky2019invariant} & 51.0 & 16.7 & 38.8 & 11.8 & 53.2 & 44.7 & 36.0  \\
DRO~\cite{sagawa2019distributionally} & 47.8 & 17.2 & 36.3 & 9.0 & 52.8 & 40.7 & 34.0  \\
Mixup~\cite{zhang2018mixup} & 55.8 & 19.2 & 46.2 & 12.8 & 58.7 & 49.2 & 40.3  \\
MLDG~\cite{li2018learning} & 59.3 & 20.3 & \textbf{48.8} & 14.0 & 61.2 & 51.2 & 42.5  \\
CORAL~\cite{sun2016deep} & 58.8 & 20.8 & 47.5 & 13.6 & 61.0 & 50.8 & 42.1  \\
MMD~\cite{li2018domain} & 54.6 & 19.6 & 44.9 & 12.6 & 59.7 & 47.5 & 39.8  \\
DANN~\cite{ganin2016domain} & 53.8 & 17.5 & 43.5 & 11.8 & 56.4  & 46.7 & 38.3\\
C-DANN~\cite{li2018deep} & 53.4 & 18.4 & 44.7 & 12.9 & 57.5  & 46.5 & 38.9\\
\hline
DSDRNet & \textbf{60.7} & \textbf{21.5} & 47.9 & \textbf{15.8} & \textbf{62.2} &\textbf{51.9} &\textbf{43.3} \\
\hline
\end{tabular}
}
\vspace{-0.1cm}
\end{table}

\section{Ablation Study}
\subsection{Impact of Different Loss}
The model involves seven different loss functions. To evaluate the performance of these loss functions on the Digits-DG dataset, we experimented with various combinations of losses to assess their impact on the model's performance. The strategy for selecting these combinations is as follows: First, the classification loss $\mathcal{L}_{\text{ce}}$, which is essential for the task, is used throughout the entire experiment. Next, we evaluate the impact of the reconstruction loss $\mathcal{L}_{\text{recon}}$. Subsequently, we test the effects of the cycle loss $\mathcal{L}_{\text{cycle}}$ and the adversarial loss $\mathcal{L}_{\text{adv}}$ on the model's performance. We then separately test the intra-instance loss $\mathcal{L}_{\text{intra}}$ and the inter-instance loss $\mathcal{L}_{\text{inter}}$. Finally, we combine $\mathcal{L}_{\text{recon}}$ and $\mathcal{L}_{\text{kl}}$ to assess the model's performance. The final results are presented in Table~\ref{tab:exp_ablat_Digits-DG} as shown in the document.

When the model only involves the $\mathcal{L}_{\text{ce}}$ loss, its performance is consistent with ERM and achieves an accuracy of 73.7\%. By adding the $\mathcal{L}_{\text{recon}}$ constraint, the performance improves by +0.5\%. This effect is similar to the impact of adding only $\mathcal{L}_{\text{cycle}}$ on the model's performance. When combining $\mathcal{L}_{\text{cycle}}$ and $\mathcal{L}_{\text{adv}}$, the performance reaches 75.4\%, which is a +1.7\% improvement over ERM. When $\mathcal{L}_{\text{intra}}$ is added, the model's performance significantly improves from 76.2\% to 77.8\%. The improvement effect of individual $\mathcal{L}_{\text{inter}}$ is similar to that of $\mathcal{L}_{\text{intra}}$, but when the intra-class $\mathcal{L}_{\text{intra}}$ and inter-class $\mathcal{L}_{\text{inter}}$ constraints are combined, the effect becomes more pronounced, directly boosting the performance to 78.8\%. By combining $\mathcal{L}_{\text{cycle}}$ and $\mathcal{L}_{\text{adv}}$, the model's final performance reaches 80.9\%. This indicates that the losses in the model are playing roles to varying degrees, with the constraints of $\mathcal{L}_{\text{intra}}$, $\mathcal{L}_{\text{inter}}$, and $\mathcal{L}_{\text{adv}}$ exerting stronger effects on the model.

\begin{table}
\centering
\caption{Ablation Study on Digits-DG dataset.}
\label{tab:exp_ablat_Digits-DG}
\resizebox{0.45 \textwidth}{!}{
\begin{tabular}{l c c c c c c |c}
%\begin{tabular}{p{1.9cm} p{1.2cm} p{1.5cm} p{1.2cm} p{1.2cm} p{1.1cm}}
\hline
$\mathcal{L}_{\text{recon}}$ & $\mathcal{L}_{\text{intra}}$ & $\mathcal{L}_{\text{inter}}$ & $\mathcal{L}_{\text{cycle}}$ & $\mathcal{L}_{\text{adv}}$ & $\mathcal{L}_{\text{ce}}$ & $\mathcal{L}_{\text{kl}}$ & Avg \\ \hline
           &            &            &            &            & \checkmark &             & 73.7 \\
\checkmark &            &            &            &            & \checkmark &             & 74.2 \\
           &            &            & \checkmark &            & \checkmark &             & 74.4 \\
           &            &            & \checkmark & \checkmark & \checkmark &             & 75.4 \\
\checkmark &            &            &            & \checkmark & \checkmark & \checkmark  & 76.2 \\
\checkmark & \checkmark &            &            &            & \checkmark & \checkmark  & 77.8 \\
\checkmark &            & \checkmark &            &            & \checkmark & \checkmark  & 77.9 \\
\checkmark & \checkmark & \checkmark &            &            & \checkmark & \checkmark  & 78.8 \\
\checkmark & \checkmark & \checkmark & \checkmark & \checkmark & \checkmark & \checkmark  & 80.9 \\
\hline
\end{tabular}}
\vspace{-0.1cm}
\end{table}

\subsection{Analyzing Loss Value Fluctuations}
\begin{figure}[!ht]
\centering
\includegraphics[width=1.0\linewidth]{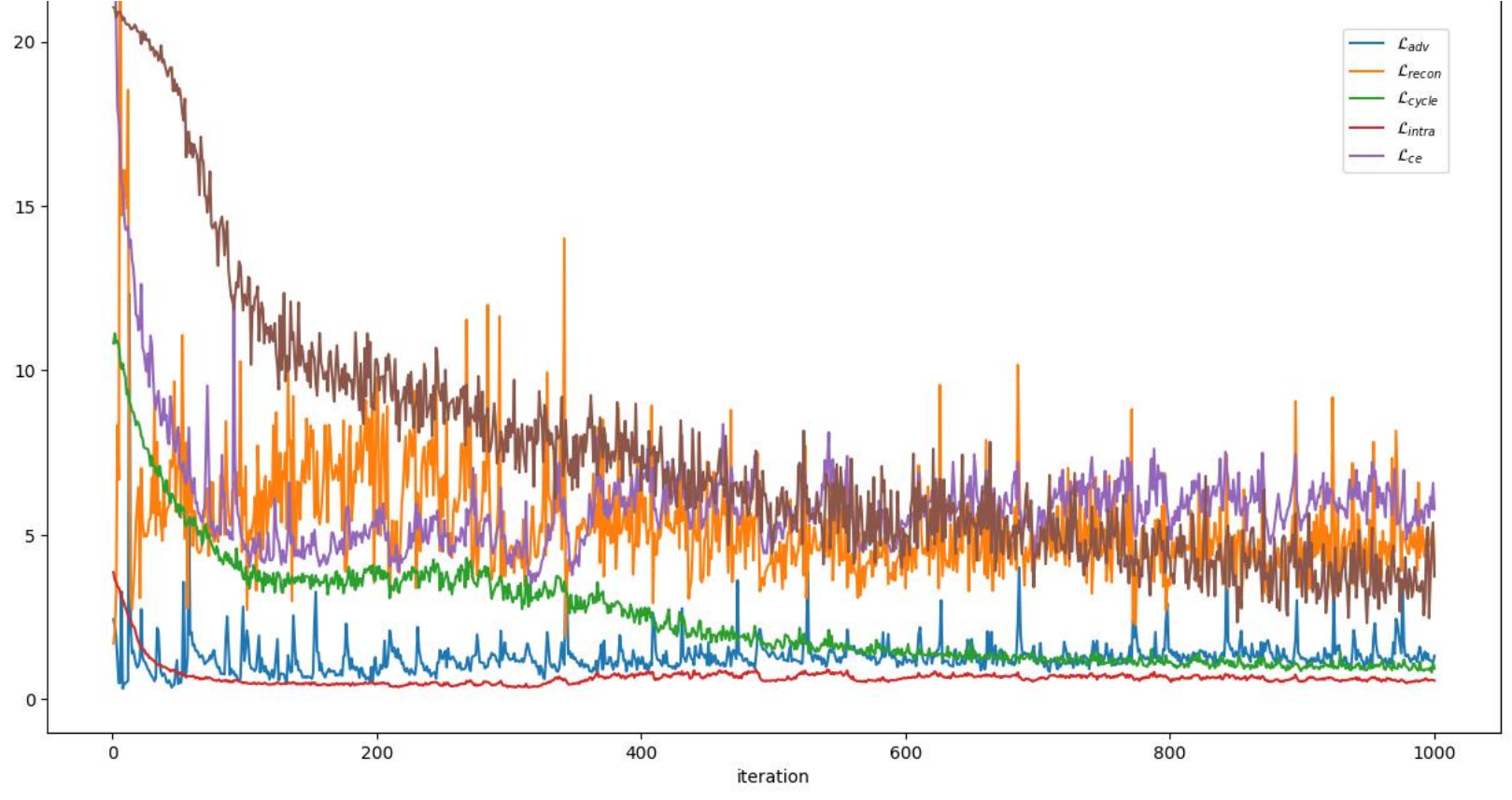}
\caption{DSDRNet's performance of $\mathcal{L}_{\text{adv}}$, $\mathcal{L}_{\text{recon}}$, $\mathcal{L}_{\text{cycle}}$, $\mathcal{L}_{\text{intra}}$ and $\mathcal{L}_{\text{ce}}$ on the MNISTM dataset.}
\label{fig:loss_lamda}
% \vspace{-0.4cm}
\end{figure}

The main loss of our DSDRNet is shown in Figure~\ref{fig:loss_lamda}, 
the details are as follows. $\mathcal{L}_{\text{adv}}$ started to oscillate decreasingly from 2.42, after 200 iterations around 0.7 and then kept oscillating between 0.5 and 2. The loss oscillated with a relatively large amplitude, it oscillated violently in the first 100 iterations and tended to stabilize at about 5 after 400 iterations. $\mathcal{L}_{\text{adv}}$ oscillated in the early stage due to relatively few training samples which led to the poor quality of the generated images. $\mathcal{L}_{\text{recon}}$ fell from the peak value of 11.12 and basically stabilized after 600 iterations and finally at around 1.5. $\mathcal{L}_{\text{cycle}}$ declined relatively fast at the beginning of training due to the relatively small amount of data and started to stabilize after 200 iterations, finally oscillating around 0.5. $\mathcal{L}_{\text{intra}}$ fell at the beginning of training from the initial 26, after 400 iterations down to 5.5, finally oscillating around 5.0. This is because, in the early oscillation, the model has not yet learned some common features because of the relatively few samples. In contrast, the later oscillation was caused by the random reading of data from the two domains where the difference in data semantics may be relatively large. As the number of training samples increases, 
$\mathcal{L}_{\text{ce}}$ decreases because the network learns more useful features with the result of a gradual improvement of the recognition rate of the samples. 

\begin{figure}[!htb]
	\centering
% 	\subfloat[Feature distribution of raw data from OfficeHome]{
		\begin{minipage}[t]{0.45\linewidth}
			\centering
			\includegraphics[width=\textwidth]{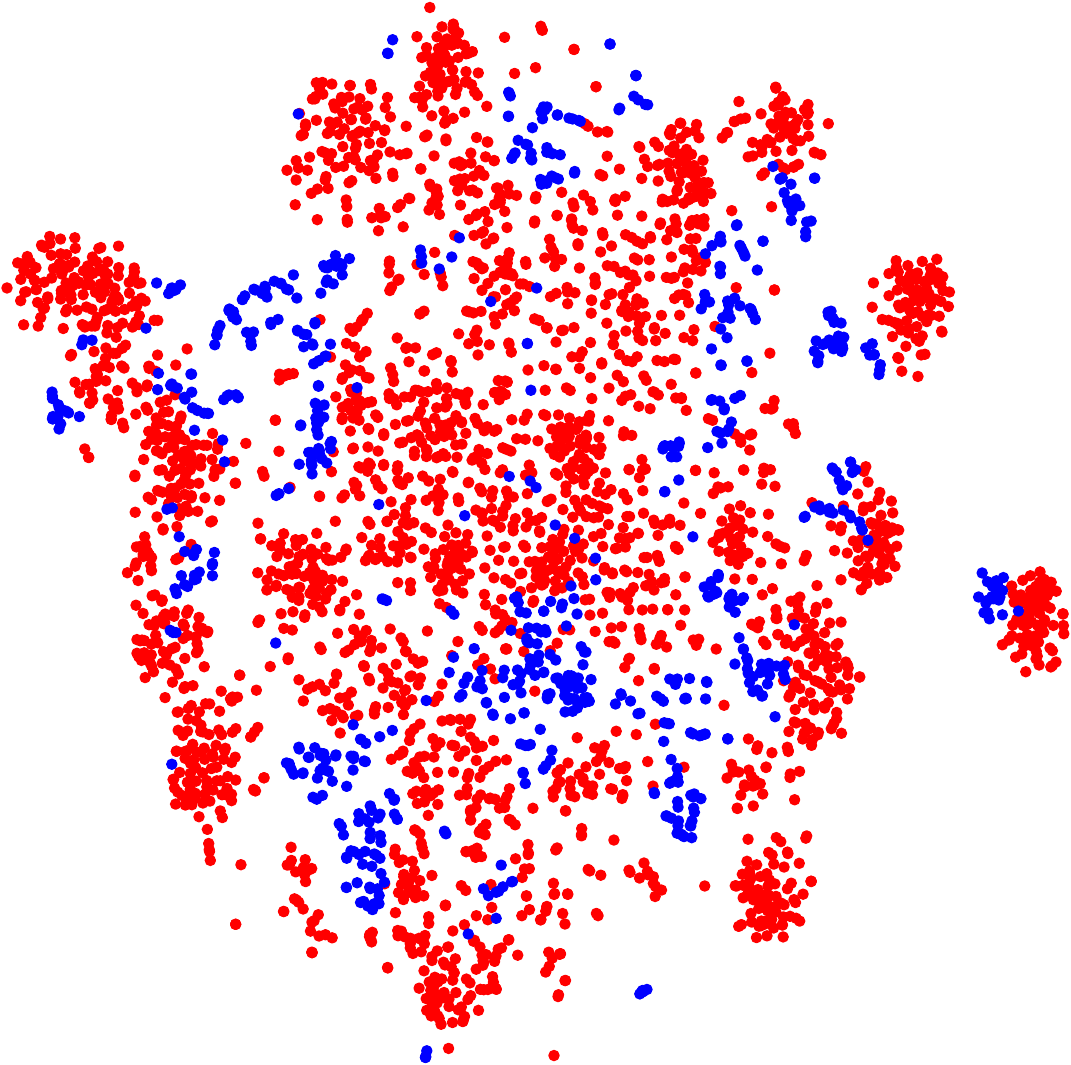}
			\label{fig::AACH}
	\end{minipage}
% 	}
% 	\subfloat[After Mix and Disentangling by DSDRNet]{
		\begin{minipage}[t]{0.45\linewidth}
			\centering
			\includegraphics[width=\textwidth]{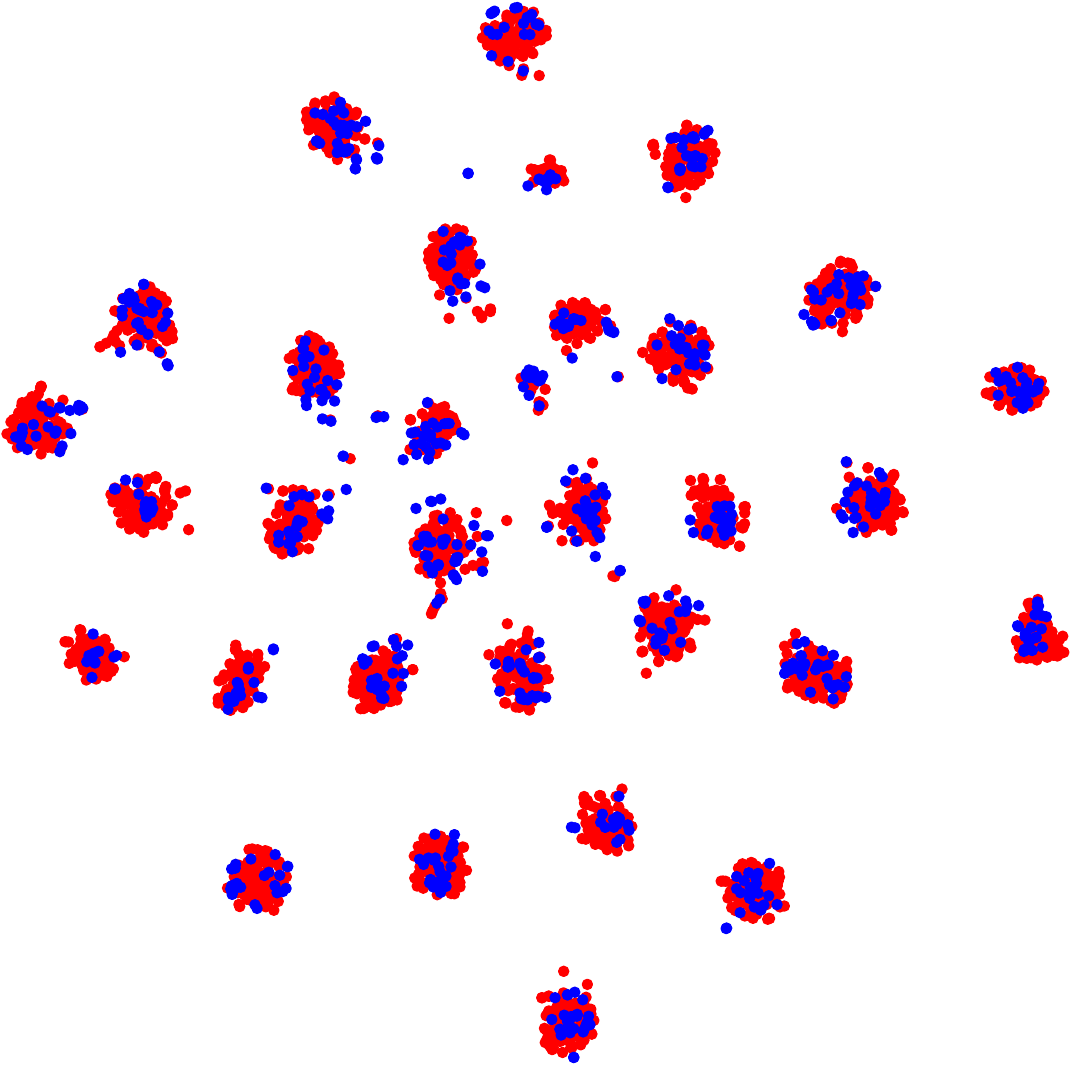}
			\label{fig::BBCH}
	\end{minipage}
% 	}
	\caption{\small t-SNE visualization of the OfficeHome. Different colors represent the different categories in the domain. (a) Feature distribution of raw data from OfficeHome. (b) Feature distribution of the sample after the processing of the DSDRNet.}
	\label{fig:officehome_tsne}
\end{figure}

\subsection{Feature Visualizations}
To further validate the performance of DSDRNet, we utilize t-SNE to visualize the OfficeHome dataset before and after processing with the model. The resulting visualization is shown in Figure~\ref{fig:officehome_tsne}. In the left image, the distribution of data from the RealWorld domain and the Product domain is mixed together, making it difficult to distinguish between different categories. However, in the right image, after processing with DSDRNet, the data from different domains are aggregated together based on their respective categories.

\section{Conclusion}\label{sec13}
In this paper, we proposed a dual-stream disentanglement and reconstruction network that achieved favorable results in four generalization tasks. However, the disentanglement and reconstruction techniques require a substantial number of loss constraints and tuning tricks to ensure the model converges to the optimal state. The interpretability of the model's performance is unknown, and the disentangled representation often consists of only one dimension that encompasses all the attribute information. The limited information in the representation hinders the method's generalization. Therefore, applying causal reasoning to disentangled representations to enhance the interpretability of the model and analyzing inter-dimensional correlations in multi-dimensional representations for improved disentanglement is a future direction worthy of attention.

\bibliographystyle{IEEEtran}
\bibliography{trans2023template}

\appendix

\subsection{Datasets}
\label{appen:datasets}
We compared DSDRNet model to state-of-the-art DG algorithms on the following tasks: digit classification(Digits-DG), image recognition(PACS~\cite{li2017deeper}, OfficeHome~\cite{venkateswara2017deep}), DomainNet~\cite{peng2019moment}). 

\textbf{Digits-DG.}$\quad$Digits-DG is a collection of four benchmarks for Digits classification, namely MNIST~\cite{lecun1998gradient}, MNISTM~\cite{ganin2015unsupervised}, Synthetic Digits~\cite{ganin2015unsupervised}, SVHN~\cite{yuval2011reading}, the main differences among these four datasets lie in image quality, background, and font style. MNIST is a classic dataset of black-and-white handwritten digits. The dataset contains images with a size of 28$\times$28 pixels and comprises 55,000 training samples and 10,000 testing samples. MNISTM is a dataset of color images featuring digits. The images in this dataset have a size of 32$\times$32 pixels and consist of 55,000 training samples and 10,000 testing samples. SVHN, which stands for street view house Numbers, is a dataset of color images showcasing house numbers. The images in this dataset are of size 32$\times$32 pixels in size and include 73,257 training samples and 26,032 testing samples. Synthetic Digits is a dataset consisting of synthetic images of English digits embedded in random backgrounds. This dataset includes 25,000 training samples.

\textbf{PACS.}$\quad$PACS~\cite{li2017deeper} is a frequently employed DG dataset consisting of 4 domains: Art Painting (2,048 images), Cartoon (2,344 images), Photo (1,670 images), and Sketch (3,929 images). This dataset encompasses seven object categories: dog, elephant, giraffe, guitar, horse, house, and person.

\textbf{OfficeHome.}$\quad$OfficeHome~\cite{venkateswara2017deep} originally introduced for domain adaptation is getting popular in the DG community. This dataset comprises approximately 15,500 images spanning 65 categories, primarily featuring Office and Home objects. Similar to PACS, it is divided into four domains: Artistic, Clipart, Product, and Real World, each of which has 65 object categories: TV, Telephone, Refrigerator, Radio, Printer, Pen, Pencil, Notebook, Mouse, Flowers, Chairs and so on.

\textbf{DomainNet.}$\quad$DomainNet~\cite{peng2019moment} is a commonly used dataset for DG tasks, consisting of 6 domains, 345 classes, and a total of 586,575 samples. The dataset is distributed across the following domains. Clipart: 48,129 samples, Infograph: 51,605 samples, Painting: 72,266 samples, Quickdraw: 172,500 samples, Real: 172,947 samples, Sketch: 69,128 samples. This dataset provides a diverse and challenging environment for DG experiments.

\subsection{Methods}
\label{appen:methods}
We compared DSDRNet to SOTA methods: DANN~\cite{ganin2016domain}, CORAL~\cite{sun2016deep}, Mixup~\cite{zhang2018mixup}, GroupDRO~\cite{sagawa2019distributionally}, RSC~\cite{huang2020self}, MMLD~\cite{matsuura2020domain}, ANDMask~\cite{parascandolo2020learning}, MixStyle~\cite{zhou2021domain}, SagNet~\cite{nam2021reducing}, CCSA~\cite{motiian2017unified}, CrossGrad~\cite{shankar2018generalizing}, DDAIG~\cite{zhou2020deep}, MMD-AAE~\cite{li2018domain}, L2A-OT~\cite{zhou2020learning}, D-SAM~\cite{d2018domain}, JiGen~\cite{carlucci2019domain}, Epi-FCR~\cite{li2019episodic}, DAEL~\cite{zhou2021domain}, IRM~\cite{arjovsky2019invariant}, DRO~\cite{sagawa2019distributionally}, MLDG~\cite{li2018learning}, MMD~\cite{li2018domain}, C-DANN~\cite{li2018deep}, and MetaReg~\cite{balaji2018metareg}. ERM as a baseline, which directly feeds all the source domain data into the model for training, without using any DG tricks. The authors' original experimental setup and results are preserved in my experiments.

\subsection{Experimental Details}
\label{appen:exper_detail}
We implement all methods with PyTorch and run experiments on two NVIDIA GeForce RTX 3090 GPUs. We use Wide\_Resnet\_28~\cite{zagoruyko2016wide} as the backbone network on the Digit-DG dataset. On the OfficeHome and PACS datasets, we utilize a pre-trained ResNet-18~\cite{he2016deep} as the backbone, which has been pre-trained on the ImageNet~\cite{russakovsky2015imagenet} dataset. However, for the DomainNet dataset, we use a ResNet-50~\cite{he2016deep} as the backbone network.

According to Domainbed~\cite{DBLP:conf/iclr/GulrajaniL21} is settings, the batch size is 64 for the Digit-DG dataset and 32 for the rest of the other datasets. The optimizer for encode $E$, discriminator $D$, generator $G$, disentangle $S$ are Adam with learning rate as $1\times 10 ^{-4}$ and weight decay as $5\times 10 ^{-4}$. We use the SGD optimizer to classifier $C$, with an initial learning rate of $5\times10^{-3}$, a momentum of 0.9, and a weight decay of $5\times10^{-4}$. In all experiments, we set the hyper-parameters as follows: $\alpha_1$ = 1, $\alpha_2$ = 1, $\alpha_3$ = 1, $\alpha_4$ = 1, $\alpha_5$ = 1, $\alpha_6$ = 8, and $\alpha_7$ = 0.5.  
\end{document}